\documentclass[nohyperref]{article}

\usepackage{microtype}
\usepackage{graphicx}
\usepackage{subfigure}
\usepackage{booktabs} 

\usepackage{hyperref}

\usepackage[accepted]{icml2022}

\usepackage{amsmath}
\usepackage{amssymb}
\usepackage{mathtools}
\usepackage{amsthm}

\usepackage[capitalize,noabbrev]{cleveref}

\theoremstyle{plain}
\newtheorem{theorem}{Theorem}

\theoremstyle{definition}
\newtheorem{definition}[theorem]{Definition}

\theoremstyle{remark}

\usepackage[textsize=tiny]{todonotes}

\icmltitlerunning{Finding the Task-Optimal Low-Bit Sub-Distribution in Deep Neural Networks}

\usepackage{url}
\usepackage{wasysym}
\usepackage{caption}
\usepackage{enumitem}
\usepackage{mathrsfs}
\usepackage{multirow}
\usepackage{pifont}
\usepackage{footnote}
\usepackage{wrapfig}
\usepackage{threeparttable}
\usepackage{array}
\usepackage{epstopdf}
\usepackage{colortbl}
\usepackage{marvosym}
\usepackage{footmisc}
\usepackage{slantsc}
\usepackage{color}
\definecolor{linecolor1}{gray}{.95} 
\definecolor{linecolor}{gray}{.895} 
\def\eg{{\it{e.g.}}}

\def\ie{{\it{i.e.}}}

\begin{document}

\twocolumn[
\icmltitle{Finding the Task-Optimal Low-Bit Sub-Distribution in Deep Neural Networks}



\icmlsetsymbol{equal}{$\dagger$}

\begin{icmlauthorlist}
\icmlauthor{Runpei Dong}{xjtu,equal}
\icmlauthor{Zhanhong Tan}{thu,equal}
\icmlauthor{Mengdi Wu}{thu}
\icmlauthor{Linfeng Zhang}{thu}
\icmlauthor{Kaisheng Ma}{thu}
\end{icmlauthorlist}

\icmlaffiliation{xjtu}{Xi'an Jiaotong University}
\icmlaffiliation{thu}{Tsinghua University}

\icmlcorrespondingauthor{Kaisheng Ma}{kaisheng@mail.tsinghua.edu.cn}

\icmlkeywords{Machine Learning, ICML}

\vskip 0.3in
]



\printAffiliationsAndNotice{\icmlEqualContribution} 

\begin{abstract}%
Quantized neural networks typically require smaller memory footprints and lower computation complexity, which is crucial for efficient deployment.
However, quantization inevitably leads to a distribution divergence from the original network, which generally degrades the performance.
To tackle this issue, massive efforts have been made, but most existing approaches lack statistical considerations and depend on several manual configurations.
In this paper, we present an adaptive-mapping quantization method to learn an optimal latent sub-distribution that is inherent within models and smoothly approximated with a concrete Gaussian Mixture (GM).
In particular, the network weights are projected in compliance with the GM-approximated sub-distribution. This sub-distribution evolves along with the weight update in a co-tuning schema guided by the direct task-objective optimization.
Sufficient experiments on image classification and object detection over various modern architectures demonstrate the effectiveness, generalization property, and transferability of the proposed method.
Besides, an efficient deployment flow for the mobile CPU is developed, achieving up to 7.46$\times$ inference acceleration on an octa-core ARM CPU. 
Our codes have been publicly released
at \url{https://github.com/RunpeiDong/DGMS}.
\end{abstract}


\begin{figure}[t]
  \begin{center}
    \includegraphics[width=\columnwidth]{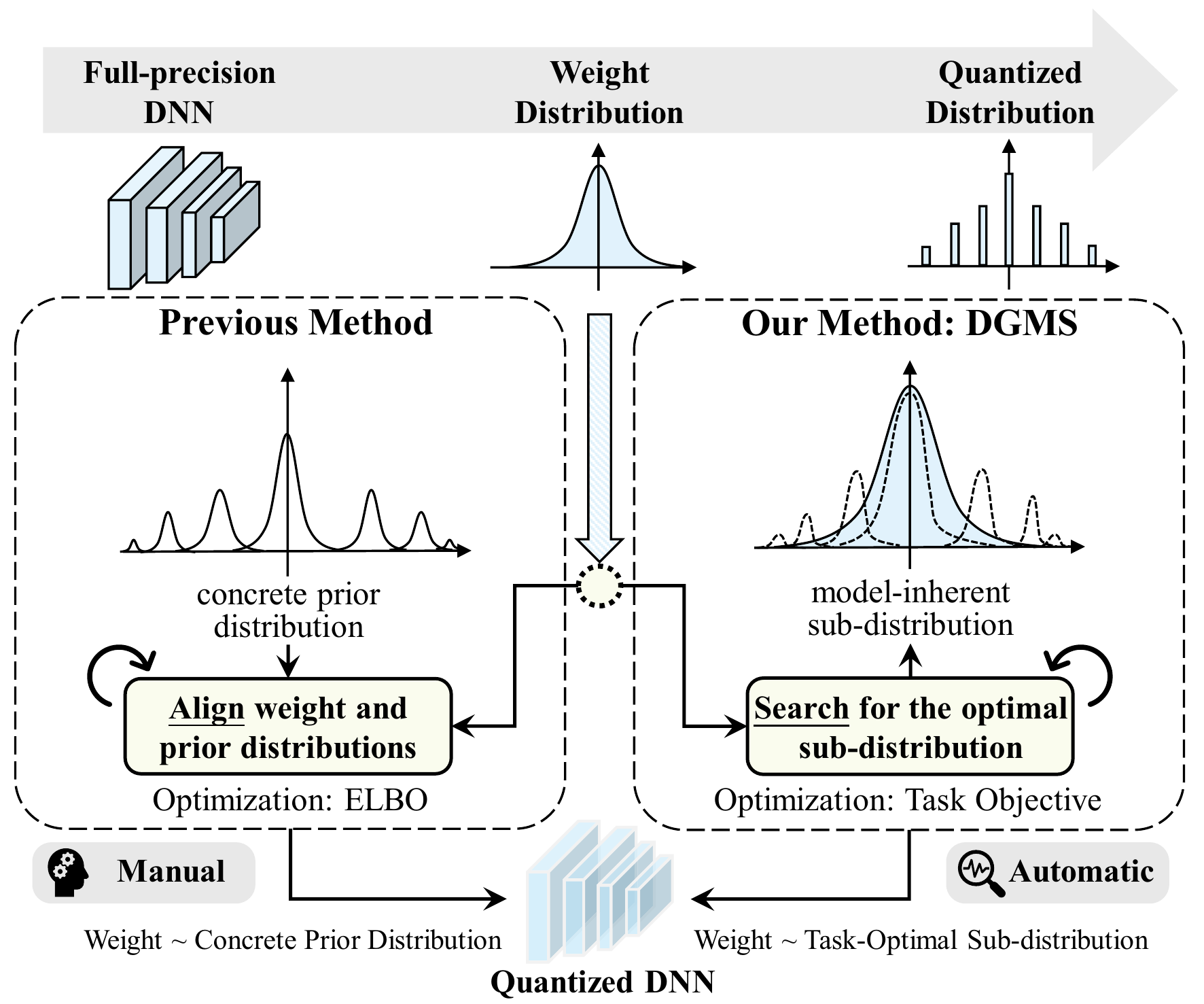}
  \end{center}
  \vspace{-6pt}
  \caption{
  Comparison between the previous methods and our proposed DGMS. 
  The training objective of the previous methods is ELBO while it is the task-objective only for DGMS, and the previous methods depend on manual configurations while our DGMS is fully automated.
  }\label{fig:concept}
\end{figure}
\vspace{-12pt}
\section{Introduction}\label{sec:intro}
Deep neural networks (DNNs) have become the \textit{de-facto} method in plentiful applications, including computer vision~\cite{ResNet, MaskRCNNPAMI}, natural language processing~\cite{BERT, BERTTheus}, speech recognition~\cite{Speech}, and robotics~\cite{RobiticDet, RobiticDepth}.
Although DNNs are pushing the limits of generalization performance across various areas, the growing memory and computation severely hinder the practical deployment.
Abundant acceleration and compression methods have been proposed to tackle this issue, such as pruning~\cite{OptimalDamage, DC}, weight quantization~\cite{WeightDiscretization, XNORNet}, knowledge distillation~\cite{TemperatureSoftmax}, efficient model design~\cite{ShuffleV2, GhostNet, MGNet}, and even neural architecture search~\cite{TransferArch}.

Among the aforementioned schemes, weight quantization, as one of the mainstream techniques, aims to eliminate the representative redundancy via shortened bit-width.
Generically, quantization can be represented as a projection $Q: \mathcal{W} \in \mathbb R \rightarrow \mathcal{Q}=\{q_0, q_1,\dots,q_K\}$, where $\mathcal{W} \sim \mathcal{D}_{\mathcal{W}}$ is the real-valued preimage while $\mathcal{Q}\sim \mathcal{D}_{\mathcal{Q}}$ denotes the compressed and discrete representation. The categorical distribution $\mathcal{D}_{\mathcal{Q}}$ can be viewed as a factorized Dirac distribution for the expected precision, which inherently suffers a divergence from the preimage $\mathcal{D}_{\mathcal{W}}$.
Consequently, the weight deviation after low-bit quantization generally leads to significant performance degradation.

To address the problem mentioned above, one typical solution is \textit{straight through estimator} (STE)~\cite{STE} that simulates and encourages the network discretization via modifying the backpropagation~\cite{QCNN,STEQuantization,PACT,GaussianQ}.
However, most of these approaches depend on the fixed configurations (\eg, centroids~\cite{BNN, INQ, WeightDiscretization} and stepsize~\cite{BinaryConnect, QuantizationDL}) and it is tough to find the optimal solution.
In addition, the global statistical information is not fully utilized,
and thus the impact of numerical state transitions is omitted~\cite{PACT}.
Accordingly, a feasible solution is to manually set a concrete prior distribution $\mathcal{D}_{\mathcal{P}}$ that smoothly approximates the factorized Dirac posterior $\mathcal{D}_{\mathcal{Q}}$ through variational learning and reparametrization~\cite{GumbelSoftmax, ConcreteDistribution, BayesianCompression, SWS}.
As shown in Fig.~\ref{fig:concept}, this method aims at minimizing the KL divergence $D_{\rm{KL}}(\mathcal{D}_{\mathcal{P}}, \mathcal{D}_{\mathcal{Q}})$ during learning the task objective, which can be achieved directly by optimizing the \textit{evidence lower bound} (ELBO)~\cite{MDLBayesian, PracticalVL}.

However, despite the fact that the differentiable learning with statistical information is adopted, it is still difficult to reach the ELBO optimum due to the large gradient variance and local optima trap.
As has been pointed out by~\citeauthor{ModelCompression}, there exists a gap between the prior and quantized distributions resulting from the separation of approximating the distribution and quantizing the weights in a two-step scheme.
Besides, direct transfer of a manual-configured prior from one domain to another seems unappealing, since it generally requires statistical alignment training.

In this paper, we introduce \textbf{D}ifferentiable \textbf{G}aussian \textbf{M}ixture Weight \textbf{S}haring (\textbf{DGMS}), a novel adaptive-mapping quantization method that statistically searches for the optimal low-bit model with a latent and representative sub-distribution.
Compared to prior works, the superiority of DGMS benefits from three aspects:
1) DGMS relaxes the Dirac posterior into a concrete GM approximation to estimate the optimal sub-distribution that is model-inherent but not user-specified.
2) DGMS directly minimizes the task loss via weight and sub-distribution co-tuning for task-optimal quantization.
3) We further leverage the temperature-based softmax reparametrization~\cite{TemperatureSoftmax, GumbelSoftmax} to narrow the gap between training and inference.

In summary, the main contributions of this paper can be listed as follows:
\begin{itemize}
    \item
    We propose a novel quantization method DGMS that statistically explores the optimal latent low-bit sub-distribution without handcrafted settings. Distributions and weights are trainable in a self-adaptive and end-to-end fashion. 
    
    \item
    The promising transfer ability across four domains reveals the domain-invariant character, indicating the model-inherence of the found sub-distribution.
    \item
    Extensive experiments with several classical and lightweight DNNs demonstrate the remarkable compression performance and generalizability of DGMS on both image classification and object detection.
    \item
    We also evaluate DGMS-quantized low-bit models on the mobile CPU with an efficient deployment flow that maintains the parallelism and optimizes the cache access. The results show 1.66$\times$ $\sim$ 7.46$\times$ speedup compared to the full-precision models.
\end{itemize}
\section{Related Works}\label{sec:related_work}
It has been acknowledged that DNNs are heavily over-parameterized with significant redundancy~\cite{DNNRedundancy}, and a wide variety of approaches have been proposed for the compact representation.
The straightforward solution is to design a lightweight model with handcrafted efficient blocks,
such as MobileNet~\cite{MBV1, MBV2, MBV3}, ShuffleNet~\cite{ShuffleV1, ShuffleV2} and GhostNet~\cite{GhostNet}, or with Neural Architecture Search (NAS)~\cite{TransferArch, EfficientNet, SNAS}. 
Another attractive direction is to eliminate the inherent redundancy in existing architectures by using Knowledge Distillation (KD)~\cite{TemperatureSoftmax,CompressionKD,SelfKD}, pruning~\cite{Removal, Han2015, DC}, or quantization~\cite{WeightDiscretization, XNORNet}.

In particular, quantization is widely recognized to be an efficient manner for DNN deployment on commercial products~\cite{huawei, tesla, ALi, MTK} profiting from the natural regularity for processing.
Quantization techniques can be basically categorized into two groups: 1) discontinuous-mapping quantization based on rounding operation, and 2) continuous-mapping quantization based on adaptive reparameterization.

\paragraph{Discontinuous-Mapping Quantization}
One line of works employs stochastic rounding~\cite{QuantizationDL, STEQuantization, Integers}, where the mapping decisions are stochastically made among quantization intervals. Particularly, RQ~\cite{RelaxedQ} is similar to our method but it employs this stochastic rounding operation for fixed quantization grids, the distribution assumption is made upon the input signal noise while we directly model the weights.
In parallel, the other line of works focuses on deterministic rounding to snap the full-precision weights to the closest quantized centroids towards binary~\cite{BinaryConnect}, ternary~\cite{DBLP:journals/tnn/MarchesiOPU93}, or power-of-two quantization~\cite{DBLP:journals/tsp/TangK93}.
Starting with the typical pioneering studies~\cite{BinaryNet, BinaryConnect}, many researchers adopt STE~\cite{STE} to tackle the non-differentiability issue caused by the rounding operation~\cite{STEQuantization, XNORNet}.
However, the pseudo-gradients may lead to an inaccurate optimization direction, and statistical information is often ignored.
Meanwhile, many methods~\cite{LQNets, ALQ,DBLP:conf/iclr/XuYLOCWZ18} have been proposed to explore Multi-Bit Quantization (MBQ)~\cite{MBQ}.
Specifically, DMBQ~\cite{DMBQ} also focuses on distribution following~\cite{ACIQ}, which employs a fixed Laplase prior to reformulate the quantization error for optimization.
Recently, KD~\cite{KDQ,CompressionKDQ} and Reinforcement Learning (RL)~\cite{ReLeQ,HAQIJCV} are introduced for better efficiency-accuracy trade-off, generally requiring indirect efforts and expertise.
\vspace{-0.8cm}

\paragraph{Continuous-Mapping Quantization}\label{sec:marker1}
This solution is able to utilize global statistical information.
Several works manually define a concrete prior distribution for approximating the quantized categorical distribution via variational learning~\cite{BayesianCompression, SWS, LearningDiscrete, EBP} or Markov Chain Monte Carlo (MCMC)~\cite{DPBNN}.
Though allowing for the differentiable learning,
the training is confronted with the convergence difficulty or large memory footprint for MCMC sampling~\cite{HMCBP}. 
Also, similar to STE-based methods, a heuristic manual prior is needed.
Our method is closely related to this continuous-mapping scheme but different from previous works. 
Our proposed DGMS automatically searches for a model-inherent sub-distribution without manual configuration, and the optimization directly targets the task-objective (see the illustration in Fig.~\ref{fig:concept} and discussions in Appendix~\ref{sec:mdl}).

To avoid the discussed issues, the primary objective of this paper is to peer inside the model-inherent statistical information for quantization in an end-to-end differentiable manner.
Our inspiration comes from the recently proposed ``Lottery Ticket Hypothesis'' in the pruning literature~\cite{LotteryTicket, LMCLottery}, which proves that there exists an optimal \textit{sub-network} for a randomly initialized DNN.
For quantization, we aim to explore whether there exists an optimal latent \textit{sub-distribution} for a given full-precision DNN that can be tuned to match the performance of the original network.

\section{Methodology}
In this section, we first present a reformulation of conventional network quantization and then introduce our proposed DGMS quantization method for the optimal sub-distribution searching.
Finally, we describe an efficient mobile deployment flow named Q-SIMD for DGMS.  
\subsection{Preliminaries: Quantization Reformulation with Auxiliary Indicator}

As stated in Sec. \ref{sec:intro}, a network quantizer $Q$ maps the continuous preimage $x\in \mathbb{R}^{M}$ to a discrete finite set $\mathcal{Q}$. For example, a symmetric uniform quantizer $Q_U$ maps given inputs in a linear fashion:
\begin{equation}\label{equ:uniformq}
    Q_U(\mathcal{W}) = 
    \mathrm{sign}(\mathcal{W}) \odot
    \begin{cases}
    \Delta \left\lfloor \frac{ \vert \mathcal{W} \vert}{\Delta} + \frac{1}{2} \right\rfloor, & \vert \mathcal{W} \vert \leq \max( \mathcal{Q}_U)\\
    \max( {\mathcal{Q}}_U), & \vert \mathcal{W} \vert > \max({\mathcal{Q}}_U)
    \end{cases}
\end{equation}
where $\odot$ denotes the Hadamard product, $\Delta$ is the predetermined stepsize and $\mathcal{Q}_U = \{0, \pm k\Delta, k=1,\dots, K\}$ is the uniform centroids.

A common problem is that this discontinuous-mapping scheme may limit the representational capacity of quantization and the nearly zero derivatives of Eqn.~(\ref{equ:uniformq}) are generally useless.
Hence, we employ an adaptive quantization set $\mathcal{Q}_A = \{\mu_0, \mu_1, \dots, \mu_K\}$ where $\mu_0=0$ and $\mu_1, \dots, \mu_K$ are adaptive values. 
Besides, an auxiliary indicator $\mathcal{I}^Q: \mathbb{R}^M \rightarrow \{0,1\}^{K\times M}$ is introduced for $\mathcal{W}$ given any condition $\mathcal{A}$ (\ie, $\mathcal{I}^Q(\mathcal{W})=1$ if $\mathcal{A}$ holds and zero otherwise), which has been widely used for pruning~\citep{Piggyback, SNIP}:
\begin{equation} \label{equ:quantizationmask}
\begin{aligned}
    \mathcal{I}^Q_k(\mathcal{W}) = \begin{cases}
        1, & \text{if}~\mathcal{W}~ \text{is quantized to}~\mu_k\\
        0, & \text{otherwise}
    \end{cases},\\\text{for}~k = 0,1, \dots K.
\end{aligned}
\end{equation}
The quantization procedure can be reformulated as follows:
\begin{equation}\label{equ:quantizationprocess}
    Q_A(\mathcal{W}) = \sum_{k=0}^{K} \mu_k \odot \mathcal{I}^Q_k(\mathcal{W}).
\end{equation}
Here, $Q_A(\mathcal{W}) \in \mathcal{Q}_A$ is the quantized low-bit representation which is defined as a one-hot linear combination of $\mathcal{Q}_A$ in Eqn.~(\ref{equ:quantizationprocess}).
With the foregoing reformulations, we further regard the auxiliary indicator as a probability-based decision variable to relax the quantization operations and serve for the end-to-end differentiable training.

\subsection{DGMS: Differentiable Gaussian Mixture Weight Sharing}\label{sec:DGMS}
First of all, DGMS aims to search for a model-inherent sub-distribution that is defined as follows:

\begin{definition}\label{def:subdistribution}
Given the preimage $\mathcal{W}\sim \mathcal{D}_{\mathcal{W}}$ and quantized data $\mathcal{Q}\sim \mathcal{D}_{\mathcal{Q}}$, the sub-distribution $\mathcal{D}_{\mathcal{S}}$ is defined as an estimation for $\mathcal{D}_{\mathcal{W}}$ (\textit{i.e.}, $\mathcal{D}_{\mathcal{S}} \approxeq \mathcal{D}_{\mathcal{W}}$, where $\approxeq$ denotes approximate equivalence), and under a parameter limitation $\tau$, $\mathcal{D}_{\mathcal{S}}$ is approximately equivalent to $\mathcal{D}_{\mathcal{Q}}$ (\textit{i.e.}, $\mathop{\lim}\limits_{\tau \rightarrow 0} \mathcal{D}_{\mathcal{S}} \approxeq \mathcal{D}_{\mathcal{Q}}$).
\end{definition}
As defined above, the sub-distribution serves as a distributional bridge between the full-precision and quantized representations, which is controlled by a hyper-parameter $\tau$ introduced later. In order to better understand Definition~\ref{def:subdistribution}, we provide the theoretical discussions in Appendix ~\ref{sec:sub-ditribution}.

\subsubsection{Gaussian Mixture Sub-Distribution Approximation}
There are a few approaches to generate the sub-distribution $\mathcal{D}_{\mathcal{S}}$ satisfying the first condition $\mathcal{D}_{\mathcal{S}} \approxeq \mathcal{D}_{\mathcal{W}}$ in Definition~\ref{def:subdistribution}, \eg, analyzing the statistical property with maximum likelihood estimation (MLE) based on a parameterized approximation.
Empirically, the pretrained weights distribute like a bell-shaped Gaussian or Laplacian~\citep{DC, FixedPoint}.
Thus, following~\citeauthor{HintonSWS}, we assume a mixture of $K+1$ uni-modal Gaussian components as a smooth sub-distribution approximation for the full-precision network.
This Gaussian Mixture (GM) approximation $\rm{GM}_{\vartheta}$ parameterized by $\vartheta=\{\pi_k, \mu_k, \gamma_k\}_{k=0}^{K}$ ($\pi_k$: prior attributing probability, $\mu_k$: mean, $\gamma_k$: standard deviation) is capable of forming a strong representation of arbitrary-shaped densities~\citep{GMM}, revealing the statistical knowledge within the model.
For efficiency and simplicity, we initialize the GM approximation with k-means algorithm, which can be viewed as a special case of expectation-maximization (EM) algorithm~\citep{EM}.

\subsubsection{Statistical Guided Weight Sharing}
The magnitude-based weight salience has been early explored~\cite{Han2015, DC}, demonstrating that the larger weights tend to outweigh those with small magnitudes.
Based on this observation, we naturally divide the weights into several regions using the parameterized GM.
Each region is associated with a Gaussian component, where the mean value serves as a region salience for the weight sharing, \ie the adaptive quantization set $\mathcal{Q}_A = \{\mu_0, \mu_1, \dots, \mu_{K}\}$ where $\mu_0=0$.
Similar to Eqn.~(\ref{equ:quantizationmask}), here we define a region decision indicator $\mathcal{I}^S: \mathbb{R}^M \rightarrow \{0,1\}^{K\times M}$ for all $K+1$ regions based on the GM sub-distribution approximation. 
Then, for each data point $\mathbf{w}_j$, the quantized low-bit representation $\Psi(\mathbf{w}_j; \vartheta) \sim \mathcal{D}_{\mathcal{Q}}$ is formulated as:
\begin{equation} \label{equ:hard_mask}
\begin{aligned}
    \mathcal{I}_k^S(\mathbf{w}_j; \vartheta) =  
    \begin{cases}
    1,  & \textnormal{if}~\mathop{\arg\max}\limits_{i} \varphi(\mathbf{w}_j, i) = k\\
    0,  & \textnormal{otherwise}
    \end{cases},\\
    \text{for}~k = 0,1,\dots,K;
\end{aligned}
\end{equation}
\begin{equation}\label{equ:hard_quantization}
    \Psi(\mathbf{w}_j;\vartheta) = 
        \sum_{k=0}^K  \mu_k \odot \mathcal{I}_k^S(\mathbf{w}_j;\vartheta),
\end{equation}
where $\varphi(\mathbf{w}_j, k) \in [0,1]$ denotes the posterior probability $p(\mathbf{w}_j \in \rm{region}~k\vert \vartheta_k)$ of $\mathbf{w}_j$ within the given region $k$ (parameterized by $\vartheta_k=\{\pi_k, \mu_k, \gamma_k\}$). This can be written as:
\begin{equation}\label{equ:rbf}
    \varphi(\mathbf{w}_j, k) = \frac{\exp \big(\pi_k
        \mathcal{N}(\mathbf{w}_j\vert\mu_k, {\gamma_k}^2)\big)}
        {\sum_{i=0}^{K} \exp \big(\pi_i
        \mathcal{N}(\mathbf{w}_j\vert\mu_i, {\gamma_i}^2) \big)}.
\end{equation}
Eqn. (\ref{equ:rbf}) is the soft-normalized region confidence according to the GM sub-distribution. 
It can also be regarded as the nearest clustering based on a weighted radial basis function (RBF) kernel~\citep{RBF} that calculates a soft symmetry distance between weights $\mathbf{w}_j$ and the corresponding region salience.
\subsubsection{Differentiable Indicator} However, recalling Eqn.~(\ref{equ:hard_mask}), the hard sampling operation $\arg\max$ makes it completely non-differentiable, making it difficult to optimize with gradient descent. 
To tackle this issue, there exist many techniques such as Gumbel softmax~\citep{GumbelSoftmax}.
In this paper, we simply adopt the temperature-based softmax~\citep{TemperatureSoftmax, ConcreteDistribution}, 
which shifts the region confidence closer to a one-hot encoding with low temperatures,
bridging the gap between the indicators during training and inference.
With the introduced technique, the distributional sampling procedure is then differentiable and can be a well estimated region confidence prediction for $\mathcal{I}_k^S(\mathbf{w}_j;\vartheta)$:
\begin{equation}\label{equ:gumbel}
\begin{aligned}
    \phi_k(\mathbf{w}_j; \vartheta, \tau) = 
    \frac{\exp \big(\varphi(\mathbf{w}_j, k)/\tau \big)}
    {\sum_{i=0}^{K} 
    \exp \big(\varphi(\mathbf{w}_j, i)/\tau \big)},\\\text{for}~k=0,1,\dots,K,
\end{aligned}
\end{equation}
where $\tau$ is set as a learnable temperature parameter that adjusts the discretization estimation level.
Further, the compressed representation $\Phi_k(\mathbf{w}_j; \vartheta, \tau) \sim \mathcal{D}_{\mathcal{S}}$ can be reformulated as follows:
\begin{equation}\label{equ_softmask}
    \Phi(\mathbf{w}_j;\vartheta, \tau) = 
        \sum_{k=0}^K  \mu_k \odot \phi_k(\mathbf{w}_j;\vartheta, \tau).
\end{equation}
\renewcommand{\algorithmicrequire}{\textbf{Input:}}
\renewcommand{\algorithmicensure}{\textbf{Output:}}
\begin{algorithm}[!ht]
	\caption{Model compression using Differentiable Gaussian Mixture Weight Sharing.}
	\label{alg:alg}
	\begin{algorithmic}[1]
	\REQUIRE Training dataset $\mathcal{D}=\{\mathcal{X}, \mathcal{Y}\}$ like ImageNet and DNN $\mathcal{F}$ of $L$ layers with full-precision weights $\mathcal{W}=\{\mathbf{w}^{\ell}\}_{\ell=1}^L$, GM component number $K+1$ and initial temperature $\{\tau^{\ell}\}_{\ell=1}^{L}$.

	\STATE \textbf{Initialization}
	\FOR {$\ell \leftarrow 1$ \textbf{to} $L$}
		\STATE $\mathcal{R} \leftarrow \left\{\mathbf{w^{\ell}}\vert\mathbf{w}^{\ell} \in \text{region}~k \right\}_{k=0}^{K}$;
		$\triangleright$ \texttt{initial region generation with k-means}
		\STATE $\min_{k=0}^{K}\left(\vert\hat{\mu}_k\vert\right) \leftarrow 0$,
		$\vartheta^{\ell} \leftarrow \left\{\hat{\mu}_k, \hat{\pi}_k\leftarrow \frac{\vert\mathcal{R}_k\vert}{\vert\mathbf{w}^{\ell}\vert}, \hat{\gamma}_k \leftarrow \sqrt{\frac{\sum^{\vert\mathbf{w}^{\ell}\vert}_{j=1} (\mathbf{w}_j^{\ell}-\hat{\mu}_k)^2}{\vert\mathbf{w}^{\ell}\vert-1}}\right\}_{k=0}^{K}$;
    \ENDFOR
	\STATE $\Theta \leftarrow  \left\{\vartheta^{\ell}, \tau^{\ell}\right\}_{\ell=1}^{L}$;
	\STATE \textbf{Training}
	\WHILE{\textnormal{not converged}}
	    \FOR{$\ell \leftarrow 1$ \textbf{to} $L$}
	        \FOR{$k \leftarrow 0$ \textbf{to} $K$}
	            \STATE $\phi_k(\mathbf{w}^{\ell}; \vartheta^{\ell}, \tau^{\ell}) \leftarrow 
				\frac{\exp \big(\varphi(\mathbf{w}^{\ell}, k)/\tau^{\ell}\big)}
				{\sum_{i=0}^{K} 
				\exp \big(\varphi(\mathbf{w}^{\ell}, i)/\tau^{\ell}\big)}$, Eqn.~(\ref{equ:rbf}) and Eqn.~(\ref{equ:gumbel});
	        \ENDFOR
	        \STATE$\Phi(\mathbf{w}^{\ell};\vartheta^{\ell}, \tau^{\ell}) \leftarrow
        \sum_{k=0}^{K}  \mu_k^{\ell} \odot \phi_k(\mathbf{w}^{\ell};\vartheta^{\ell}, \tau^{\ell})$, Eqn.~(\ref{equ_softmask});
	    \ENDFOR
	    \STATE Evaluate with task loss, \eg, Cross Entropy $\mathcal{L}_{\rm{CE}}\Big(\mathcal{F}\big(\mathcal{X}; \Phi(\mathcal{W};\Theta)\big), \mathcal{Y}\Big)$;
	    \STATE Backpropagation and update $\{\mathcal{W},\Theta\}$ with the stochastic gradient descent;
	\ENDWHILE
	\STATE\textbf{Inference}
	\STATE $\widetilde{\mathcal{W}} \leftarrow \Psi(\mathcal{W}; \Theta)$, Eqn.~(\ref{equ:hard_mask}) and Eqn.~(\ref{equ:hard_quantization});
	\STATE Output prediction $\widehat{\mathcal{Y}}$ with quantized neural network: $\widehat{\mathcal{Y}}=\mathcal{F}(\mathcal{X}; \widetilde{\mathcal{W}})$.
\end{algorithmic}
\end{algorithm}

\subsubsection{Training and Inference}
The proposed DGMS quantization can be summarized in Algorithm \ref{alg:alg}. Note that DGMS can be deployed along different granularities such as channel-wise or filter-wise. In this paper, layer-wise DGMS is explored since it is faster to train.

\subsection{Q-SIMD: Efficient Deployment on the Mobile CPU}
\label{ch:QSIMD}
DGMS not only reduces the memory footprint but also speedups the inference for deployment.
To evaluate the model acceleration benefiting from DGMS, we develop an efficient flow to deploy our quantized 4-bit networks on the mobile CPU. The deployment faces two challenges: 1) to fetch real weights with 4-bit indices while maintaining the parallelism (i.e., SIMD), and 2) to optimize the shifted cache bottleneck of activations after the weight compression.

Given that the smallest numeric data type in current ARM-v8 is \texttt{BYTE}, simply employing the 4-bit indices wastes the high half-byte for the 8-bit alignment. Therefore, we propose an 8-bit associated index that can access multiple weights at a time. For example, we use a 256-entry codebook to represent a 4-bit index set, where each entry contains a couple of weights, $(W_X, W_Y)$. The associated index fully utilizes 8 bits, and naturally, it is conducive to loading data in parallel for SIMD operations. Even though the codebook size is squared, the overhead can be ignored compared to the whole index tensor. 
Besides, after weight quantization, the memory bottleneck has turned to be the activation access. To reduce cache misses, we further place the activation-related iterations to the outer loops so that activations can be kept local in the L1 cache without data reloading. More details can be found in Appendix~\ref{sec:QSIMD}.

\section{Experiments}\label{sec:exp}
\subsection{Experimental Setup}\label{sec:setup}
\subsubsection{Datasets and Models}\label{sec:marker2}
To demonstrate the effectiveness and generalization ability of DGMS, we evaluate our method on image classification and object detection. 
Classification experiments are conducted on CIFAR-10~\citep{CIFAR} and ImageNet~\citep{ImageNet}. 
PASCAL VOC~\citep{PASCAL} dataset is used as the detection benchmark. 
We use VOC2007 plus VOC2012 \texttt{trainval} for training and evaluate on VOC2007 \texttt{test}.
We principally select a series of lightweight models to achieve further compression, which is more significant for an advanced quantization technique at present.
For CIFAR-10, we evaluate DGMS on ResNet-20, ResNet-32, ResNet-56~\citep{ResNet} and 
VGG-Small~\citep{LQNets} (a small-sized variant of VGG-Net~\citep{VGG}). 
For ImageNet, we use ResNet-18 and ResNet-50 for evaluations~\citep{ResNet}.
Besides, we evaluate DGMS on lightweight architectures including MobileNetV2~\citep{MBV2} and two NAS-based MnasNet-A1~\citep{MnasNet} and ProxylessNAS-Mobile~\citep{ProxylessNAS}.
The experiments on PASCAL VOC are performed on SSDLite, a publicly available lite version of SSD~\citep{SSD}. 
Here, we use SSDLite-MBV2 and SSDLite-MBV3 with MobileNetV2~\citep{MBV2} and MobileNetV3~\citep{MBV3} as the backbone models, respectively.
\subsubsection{Deployment}
We extend the TVM framework~\citep{TVM} to support the Q-SIMD flow introduced in Sec.~\ref{ch:QSIMD}.
Our evaluation is performed on the octa-core ARM CPU in Qualcomm 888 (Samsung S21). 
We conduct various image classification and object detection models quantized by DGMS to evaluate their runtime using Android TVM RPC tool. The baseline 
results are obtained under full-precision models using 
the primitive optimal TVM scheduling.

\begin{table}[t!]
    \centering
    \caption{Comparison across different quantization methods on CIFAR-10. \#Params: the number of non-zero model parameters, Bits: weights quantization bit-width, TOO: task-objective optimization only. Compared methods are TTQ~\cite{TernaryQ} and LQNets~\cite{LQNets}.}\label{tab:cifar}
    \resizebox{\columnwidth}{!}{
    \begin{tabular}{llcccc}
    \toprule[0.95pt]
    Model & Method & TOO & Bits & \#Params & Top-1 Acc.\\
    \midrule[0.6pt]
    \multirow{5}{*}[-1.0ex]{ResNet-20} & Our FP32 & N/A & FP32 & 0.27M & 93.00\%\\
    \cmidrule(){2-6}
    & LQNets & $\times$ & 3 & 0.27M & 92.00\%\\
    & \cellcolor{linecolor}\textbf{Ours} & \cellcolor{linecolor}\checkmark & \cellcolor{linecolor}3 & \cellcolor{linecolor}\textbf{0.20M} & \cellcolor{linecolor}\textbf{92.84\%}\\
    \cmidrule(){2-6}
    & LQNets & $\times$ & 2 & 0.27M & 91.80\%\\
    \rowcolor{linecolor}\cellcolor{white}& \textbf{Ours} & \checkmark & 2 & \textbf{0.15M} & \textbf{92.13\%}\\
    \midrule[0.6pt]
    \multirow{3}{*}[-0.5ex]{ResNet-32}
    & Our FP32 & N/A & FP32 & 0.46M & 94.27\%\\
    \cmidrule(){2-6}
    & TTQ & \checkmark & 2 & 0.46M & 92.37\%\\
    \rowcolor{linecolor}\cellcolor{white}& \textbf{Ours} & \checkmark & 2 & \textbf{0.27M} & \textbf{92.97\%}\\
    \midrule[0.6pt]
    \multirow{3}{*}[-0.5ex]{ResNet-56}
    & Our FP32 & N/A & FP32 & 0.85M & 94.61\%\\
    \cmidrule(){2-6}
    & TTQ & \checkmark & 2 & 0.85M & 93.56\%\\
    \rowcolor{linecolor}\cellcolor{white}& \textbf{Ours} & \checkmark  & 2 & \textbf{0.44M} & \textbf{93.72\%}\\
    \midrule[0.6pt]
    \multirow{5}{*}[-1.0ex]{VGG-Small}
    & Our FP32 & N/A & FP32 & 4.66M & 94.53\%\\
    \cmidrule(){2-6}
    & LQNets & $\times$ & 3 & 4.66M & 93.80\%\\
    & \cellcolor{linecolor}\textbf{Ours} & \cellcolor{linecolor}\checkmark & \cellcolor{linecolor}3 & \cellcolor{linecolor}\textbf{3.12M} & \cellcolor{linecolor}\textbf{94.46\%}\\
    \cmidrule(){2-6}
    & LQNets & $\times$ & 2 & 4.66M & 93.80\%\\
    \rowcolor{linecolor}\cellcolor{white}& \textbf{Ours} & \checkmark & 2 & \textbf{2.24M}& \textbf{94.36\%}\\
    \bottomrule[0.95pt]
    \end{tabular}
    }
\end{table}
\begin{table}[!ht]
    \centering
    \setlength\tabcolsep{2.9pt}
    \caption{Comparison across different quantization-only and joint quantization-pruning methods on ImageNet. P+Q: joint pruning-quantization methods. Compared methods are LQNets~\cite{LQNets}, AutoQ~\cite{AutoQ}, HAQ~\cite{HAQIJCV}, TQT~\cite{TQT}, BB~\cite{BB}, CLIP-Q~\cite{CLIPQ}, ANNC~\cite{ANNC}, UNIQ~\cite{UNIQ} and HAWQV3~\cite{HAWQV3}.}
    \label{tab:imagenet}
    \resizebox{\columnwidth}{!}{
    \begin{threeparttable}
    \begin{tabular}{llcccc}
    \toprule[0.95pt]
    Model & Method & P+Q & W/A & Top-1 Acc. & $\Delta$ Acc. \\
    \midrule[0.6pt]
    \multirow{12}{*}[-2.0ex]{ResNet-18} 
    & Our FP32 & $\times$ & FP32 & 69.76\% & N/A\\
    \cmidrule(){2-6}
    & LQNets & $\times$ & 4/32 & 70.00\% & +0.40\%\\
    \rowcolor{linecolor}\cellcolor{white}& \textbf{Ours} & $\times$ & 4/32 & \textbf{70.25\%} & \textbf{+0.49\%}\\
    \cmidrule(){2-6}
    & LQNets & $\times$ & 3/32 & 69.30\% & -0.30\%\\
    \rowcolor{linecolor}\cellcolor{white}& \textbf{Ours} & $\times$  & 3/32 & \textbf{69.57\%} & \textbf{-0.19\%}\\
    \cmidrule(){2-6}
    & BB\tnote{$\flat$} & \checkmark  & 4/8 & 69.60\% & -0.08\%\\
    & UNIQ & $\times$ & 4/8 & 67.02\% & -2.58\%\\
    & HAWQV3 & $\times$  & 4/8 mixed & \textbf{70.22\%} & -1.25\%\\
    \rowcolor{linecolor}\cellcolor{white}& \textbf{Ours} & $\times$ & 4/8 & 70.12\% & \textbf{+0.46\%}\\
    \cmidrule(){2-6}
    & AutoQ	& $\times$  & 4/4 & 67.30\% & -2.60\%\\
    & HAWQV3 & $\times$  & 4/4 & 68.45\% & -3.02\% \\
    \rowcolor{linecolor}\cellcolor{white}& \textbf{Ours} & $\times$ & 4/4 & \textbf{68.95\%} & \textbf{-0.81\%}\\
    \midrule[0.6pt]
    \multirow{16}{*}[-2.5ex]{ResNet-50} & Our FP32 & $\times$ & FP32 & 76.15\% & N/A\\
    \cmidrule(){2-6}
    & UNIQ & $\times$ & 4/32 & 75.09\% & -0.93\%\\
    & HAQ & $\times$ & 4/32 mixed & 76.14\% & -0.01\%\\
    \rowcolor{linecolor}\cellcolor{white}& \textbf{Ours} & $\times$ & 4/32 & \textbf{76.28\%} & \textbf{+0.13\%}\\
    \cmidrule(){2-6}
    & CLIP-Q\tnote{$\natural$} & \checkmark & 3/32 & 73.70\% & \textbf{+0.60\%}\\
    & HAQ & $\times$ & 3/32 mixed & 75.30\% & -0.85\%\\
    \rowcolor{linecolor}\cellcolor{white}& \textbf{Ours} & $\times$  & 3/32 & \textbf{75.91\%} & -0.24\%\\
    \cmidrule(){2-6}
    & HAQ & $\times$ & 2/32 mixed & 70.63\% & -5.52\%\\
    \rowcolor{linecolor}\cellcolor{white}& \textbf{Ours} & $\times$ & 2/32 & \textbf{72.88\%} & \textbf{-3.27\%}\\
    \cmidrule(){2-6}
    & UNIQ & $\times$ & 4/8 & 74.37\% & -1.65\%\\
    & TQT & $\times$  & 4/8 & 74.40\% & -1.00\%\\
    & HAWQV3 & $\times$  & 4/8 mixed & 75.39\% & -2.33\%\\
    \rowcolor{linecolor}\cellcolor{white}& \textbf{Ours} & $\times$ & 4/8 & \textbf{76.22\%} & \textbf{+0.07\%}\\
    \cmidrule(){2-6}
    & AutoQ & $\times$  & 4/4 & 72.43\% & -2.37\%\\
    & HAWQV3 & $\times$  & 4/4 & 74.24\% & -3.48\%\\
    \rowcolor{linecolor}\cellcolor{white}& \textbf{Ours} & $\times$ & 4/4 & \textbf{75.05\%} & \textbf{-1.10\%}\\
    \midrule[0.6pt]
    \multirow{4}{*}[-1.0ex]{MnasNet-A1} & Our FP32 & $\times$ & FP32 & 73.51\% & N/A\\
    \cmidrule(){2-6}
    & ANNC\tnote{$\sharp$} & \checkmark  & 4/32 & 71.80\% & -1.66\%\\
    & \cellcolor{linecolor}\textbf{Ours} & \cellcolor{linecolor}$\times$ & \cellcolor{linecolor}4/32 & \cellcolor{linecolor}\textbf{72.18\%} & \cellcolor{linecolor}\textbf{-1.33\%}\\
    \cmidrule(){2-6}
    \rowcolor{linecolor}\cellcolor{white}& \textbf{Ours} & $\times$ & 4/8 & \textbf{72.09\%} & \textbf{-1.42\%}\\
    \midrule[0.6pt]
    \multirow{4}{*}[-1.0ex]{\shortstack{ProxylessNAS\\-Mobile}} & Our FP32 & $\times$ & FP32 & 74.59\% & N/A\\
    \cmidrule(){2-6}
    & ANNC\tnote{$\sharp$} & \checkmark  & 4/32 & 72.46\% & -2.13\%\\
    & \cellcolor{linecolor}\textbf{Ours} & \cellcolor{linecolor}$\times$  & \cellcolor{linecolor}4/32 & \cellcolor{linecolor}\textbf{73.85\%} & \cellcolor{linecolor}\textbf{-0.74\%}\\
    \cmidrule(){2-6}
    & \cellcolor{linecolor}\textbf{Ours} & \cellcolor{linecolor}$\times$ & \cellcolor{linecolor}4/8 & \cellcolor{linecolor}\textbf{73.24\%} & \cellcolor{linecolor}\textbf{-1.35\%}\\
    \bottomrule[0.95pt]
    \end{tabular}
    \begin{tablenotes}
    \item[$\flat$] Result estimated from the figure plot reported in the original paper. The weight CR of BB is $8.93\times$ while for DGMS the CR is $9.41\times$.
    \item[$\natural$] The weight CR of CLIP-Q is $31.81\times$ while DGMS yields CR of $14.75\times$ and $31.37\times$ for 3-bit and 2-bit ResNet50, respectively.
    \item[$\sharp$] The MnasNet CR is $17.10\times$ with ANNC compared to $12.40\times$ with DGMS; the ProxylessNAS CR is $16.80\times$ with ANNC compared to $9.02\times$ with DGMS.
    \end{tablenotes}
    \end{threeparttable}
    }
    \vspace{-10pt}
\end{table}
\subsection{Image Classification Results}
\subsubsection{CIFAR-10 Results} 
As shown in Tab.~\ref{tab:cifar}, we compare the proposed DGMS on CIFAR-10 to LQNets~\citep{LQNets} which is trained with Quantization Error Minimization (QEM), and TTQ~\citep{TernaryQ} which is trained with task-objective optimization only (see discussions in Sec.~\ref{sec:mdl}).
From the experimental results, we observe that DGMS consistently achieves a promising model sparsity at the same bit-width with a more lightweight representation while maintaining the performance.
For example, the 2-bit ResNet-20 architecture that is extremely tiny can still achieve 44.44\% parameter reduction. 
Besides, when quantizing models under a lower-precision bit-width setting, a higher model-inherent sparsity is found by DGMS, which is derived from the decreasing number of GM components (see Sec.~\ref{sec:GM_component}). 
\begin{table*}[ht]
    \setlength\tabcolsep{2.pt}
    \centering
    \caption{Compression results for SSDLite detectors on PASCAL VOC2007.
    For each model, the first and the second row show the full model performance (\%) and our compressed detector, respectively.}
    \label{tab:voc}
    \resizebox{\textwidth}{!}{
    \small
    \begin{tabular}{lcc|c|cccccccccccccccccccccc}
    \toprule[0.95pt]
    \footnotesize Model & \footnotesize \#Params  &\footnotesize Bits&\footnotesize mAP&\footnotesize aero&\footnotesize bike&\footnotesize bird&\footnotesize boat&\footnotesize bottle&\footnotesize bus&\footnotesize car&\footnotesize cat&\footnotesize chair&\footnotesize cow&\footnotesize dog&\footnotesize horse&\footnotesize mbike&\footnotesize person&\footnotesize plant&\footnotesize sheep&\footnotesize sofa&\footnotesize table&\footnotesize train&\footnotesize tv\\
    \midrule[0.6pt]
    \multirow{2}{*}[-0.5ex]{\shortstack{SSDLite\\-MBV2}} 
    & 3.39M & 32 & 68.6
    & 69.7 & 78.2 & 63.4 & 54.8 & 35.6 & 78.8 & 74.4 & 82.0 & 53.8 & 61.9 & 78.5 & 82.2 & 80.6 & 71.8 & 42.8 & 62.6 & 78.4 & 73.7 & 83.3 & 65.5\\
    \cmidrule(){2-24}
    & \cellcolor{linecolor}2.68M & \cellcolor{linecolor}4 & \cellcolor{linecolor}67.9
    & \cellcolor{linecolor}69.4 & \cellcolor{linecolor}78.5 & \cellcolor{linecolor}63.6 & \cellcolor{linecolor}54.6 & \cellcolor{linecolor}34.2 & \cellcolor{linecolor}77.6 & \cellcolor{linecolor}73.1 & \cellcolor{linecolor}82.4 & \cellcolor{linecolor}53.1 & \cellcolor{linecolor}61.1 & \cellcolor{linecolor}76.3 & \cellcolor{linecolor}81.3 & \cellcolor{linecolor}81.3 & \cellcolor{linecolor}70.7 & \cellcolor{linecolor}41.0 & \cellcolor{linecolor}62.3 & \cellcolor{linecolor}78.8 & \cellcolor{linecolor}72.6 & \cellcolor{linecolor}81.1 & \cellcolor{linecolor}65.0\\
    \midrule[0.6pt]
    \multirow{2}{*}[-0.5ex]{\shortstack{SSDLite\\-MBV3}}
    & 2.39M & 32 & 67.2
    & 66.0 & 79.7 & 58.7 & 54.9 & 36.9 & 79.3 & 73.0 & 83.3 & 51.1 & 62.8 & 77.7 & 81.5 & 77.1 & 70.0 & 39.0 & 58.9 & 74.6 & 72.0 & 82.2 & 64.8\\
    \cmidrule(){2-24}
    &  \cellcolor{linecolor}1.26M & \cellcolor{linecolor}4 & \cellcolor{linecolor}65.7
    & \cellcolor{linecolor}65.5 & \cellcolor{linecolor}77.9 & \cellcolor{linecolor}55.5 & \cellcolor{linecolor}54.5 & \cellcolor{linecolor}34.3 & \cellcolor{linecolor}77.7 & \cellcolor{linecolor}72.1 & \cellcolor{linecolor}83.8 & \cellcolor{linecolor}48.7 & \cellcolor{linecolor}59.6 & \cellcolor{linecolor}76.2 & \cellcolor{linecolor}80.2 & \cellcolor{linecolor}75.8 & \cellcolor{linecolor}69.2 & \cellcolor{linecolor}37.2 & \cellcolor{linecolor}57.0 & \cellcolor{linecolor}72.3 & \cellcolor{linecolor}72.5 & \cellcolor{linecolor}80.8 & \cellcolor{linecolor}62.5\\
    \bottomrule[0.95pt]
\end{tabular}
}
\end{table*}

\subsubsection{ImageNet Results}
In Tab.~\ref{tab:imagenet}, we show the classification performance of compressed models on the ImageNet benchmark, compared to several strong baselines including LQNets~\citep{LQNets}, HAQ~\citep{HAQIJCV}, TQT~\citep{TQT}, AutoQ~\citep{AutoQ}, UNIQ~\citep{UNIQ}, HAWQV3~\citep{HAWQV3} and joint pruning-quantization methods including BB~\citep{BB}, CLIP-Q~\citep{CLIPQ} and ANNC~\citep{ANNC}.
Note that we adopt full-precision models from torchvision 0.7.0 and torch hub.
Since this work focuses on weights quantization, the weights are quantized with DGMS and activations are quantized in a post-training fashion following~\citeauthor{BRECQ}.
Tab.~\ref{tab:imagenet} shows that for the 4-bit quantization setting, our DGMS-quantized ResNet-18 and ResNet-50 achieve even higher classification accuracy than the full-precision models. Even though lightweight NAS-based models are challenging for compression, the efficient MnasNet-A1 and ProxylessNAS-Mobile quantized by our DGMS lead to a negligible accuracy loss.
As a result, DGMS presents a competitive or better classification accuracy with consistently decent CRs in comparison to both quantization-only and joint pruning-quantization model compression methods.

\subsection{Object Detection Results}
To demonstrate the generalization ability of our proposed DGMS, we quantize the lightweight detectors SSDLite-MBV2 and SSDLite-MBV3 to 4-bit models on PASCAL VOC benchmark.
Note that our method can be deployed on these detectors directly with no additional modification efforts.
As displayed in Tab.~\ref{tab:voc}, we observe 10.12$\times$ and 15.17$\times$ compression rates respectively for SSD-MBV2 and -MBV3 while retaining the detection accuracy. 
For some objects like \textit{motor-bike} (mbike), our quantized detectors can achieve more accurate perception and we argue that the degradation comes from the lack of training samples and the intrinsic detection difficulty for tiny objects. 
\subsection{Deployment Results on the Mobile CPU}
Tab.~\ref{tab:compiler} compares the inference performance of the Q-SIMD flow with primitive TVM scheduling on the mobile CPU. It is observed that a 4.34-to-7.46$\times$ speedup is achieved for ResNet-18 and ResNet-50, respectively.
Lightweight DNNs have scant room because separable convolutions are widely used in models with less weight reuse probability and reduced computation. 
Despite this, a 1.66-to-1.74$\times$ speedup is achieved in three lightweight models.
\begin{table}[h!]
    \centering
    \setlength\tabcolsep{2pt}
    \caption{Deployment results on the mobile CPU of Qualcomn 888 for the ImageNet and PASCAL VOC2007 runtime (ms). Baseline: full-precision models based on primitive TVM~\cite{TVM}. DGMS: 4-bit quantized models based on Q-SIMD deployment.}\label{tab:compiler}
    \resizebox{\columnwidth}{!}{
    \begin{threeparttable}
    \begin{tabular}{lccccc}
    \toprule[0.95pt]
    Model &ResNet-18 & ResNet-50 & MnasNet-A1 & ProxylessNAS & SSDLite\tnote{$\diamondsuit$}\\
    \midrule[0.6pt]
    Baseline & 67.09 ms & 148.56 ms& 18.27 ms  & 22.17 ms  & 43.95 ms \\
    \rowcolor{linecolor}\textbf{DGMS} & \textbf{8.99 ms} & \textbf{34.26 ms} & \textbf{10.49 ms} & \textbf{13.34 ms} & \textbf{25.60 ms}\\
    \midrule[0.6pt]
     Speedup & 7.46$\times$ & 4.34$\times$ & 1.74$\times$& 1.66$\times$& 1.72$\times$\\
    \bottomrule[0.95pt]
    \end{tabular}
    \begin{tablenotes}
    \item[$\diamondsuit$] SSDLite-MBV2 with MobileNetV2 as the backbone model.
    \end{tablenotes}
    \end{threeparttable}
    }
\end{table}

\subsection{Ablation Study}
\subsubsection{GM Component Number}\label{sec:GM_component}
\begin{figure}[!h]
  \begin{center}
    \includegraphics[width=0.78\columnwidth]{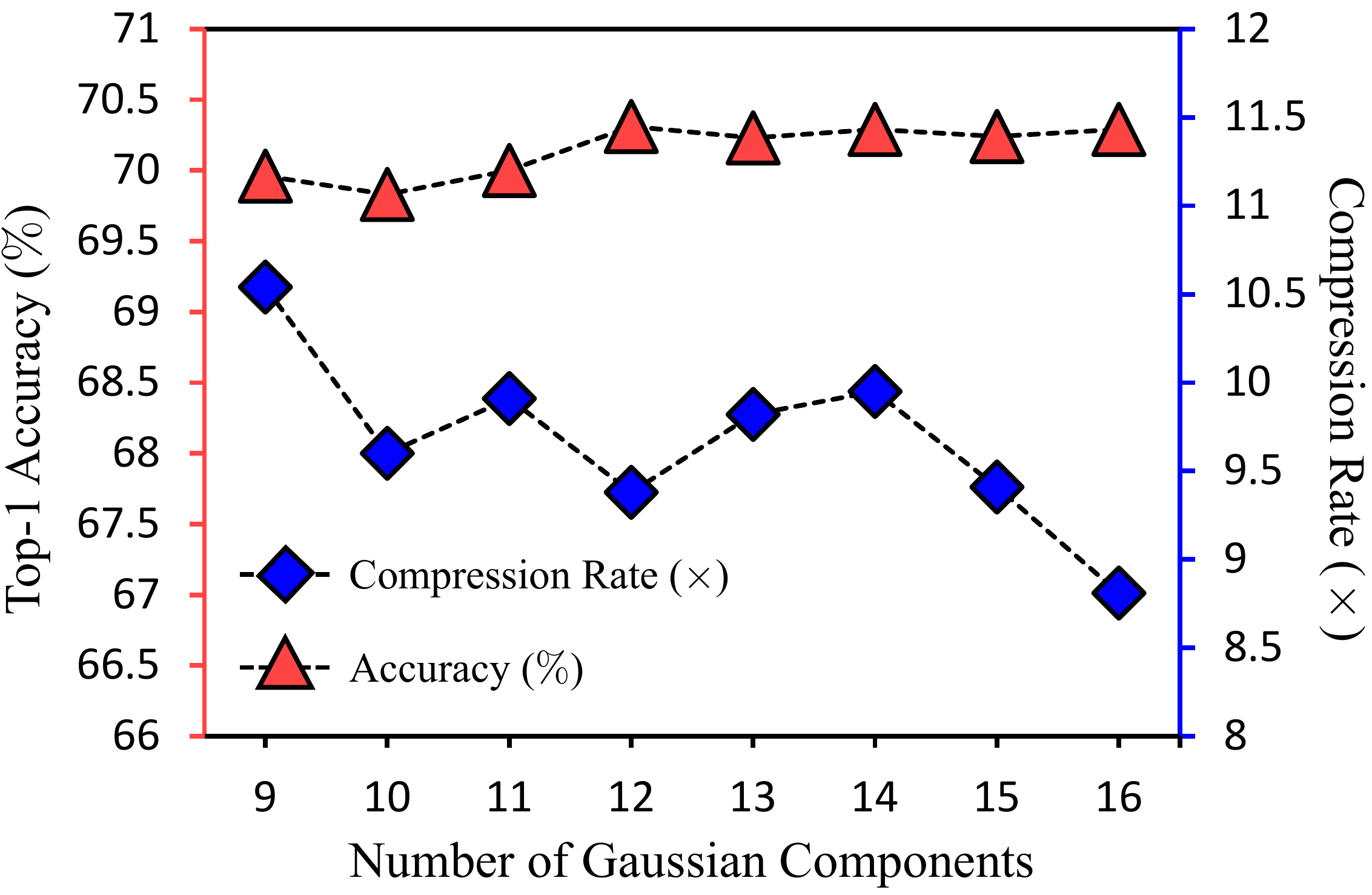}
  \end{center}
  \caption{ResNet-18 ImageNet performance and CR with varying numbers of GM components.}\label{fig:ablation}
\end{figure}

Fig.~\ref{fig:ablation} plots the Top-1 accuracy (\%) of 4-bit quantized ResNet-18 on ImageNet. The corresponding compression rates with varying numbers of GM components (\ie, varying $K$).
The figure exhibits that the compression rates slightly decline when the accuracy boosts for large values of $K$.
It indicates that GM with fewer components is more compact with higher sparsity, which spontaneously results from the lower degree of representation freedom.


\section{Discussions}\label{sec:discussion}
\begin{figure*}[!ht]
\centering
    \includegraphics[width=\linewidth]{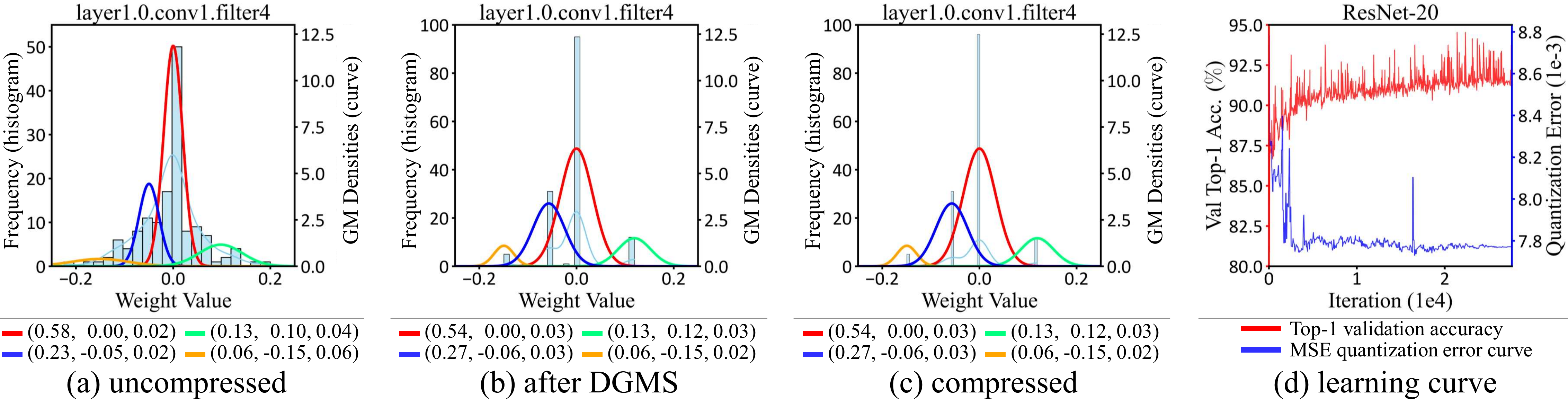}
   \caption{(a)-(c) Weight distribution and the Gaussian densities of GM. The legend (x, y, z) represents $(\pi=x, \mu=y, \gamma=z)$. (d) Top-1 validation accuracy (\%) and MSE quantization error curve.}
\label{fig:distribution}
\end{figure*}
\begin{table*}[!ht]
\setlength\tabcolsep{5.875pt}
        \centering
        \caption{Top-1 accuracy of 4-bit quantized ResNet-18 across domains. DGMS without (w/o) transfer is the vanilla DGMS on the Target domain using GM initialization method introduced in Algorithm~\ref{alg:alg}, and
        with (w/) transfer denotes GM initialization with sub-distribution found in the Source domain (Source$\rightarrow$Target domain sub-distribution transfer).}
        \label{tab:domain_study}
        \resizebox{\textwidth}{!}{
        \begin{tabular}{llcccccccccccc}
        \toprule[0.95pt]
             \multicolumn{2}{l}{Target} & \multicolumn{3}{c}{\textbf{Im}a\textbf{g}eNet} & \multicolumn{3}{c}{\textbf{CUB}200-2011} & \multicolumn{3}{c}{Stanford \textbf{Cars}} & \multicolumn{3}{c}{FGVC \textbf{Air}craft}\\
             \midrule[0.6pt]
             \multicolumn{2}{l}{FP32 Full-Model} & \multicolumn{3}{c}{69.76\%} & \multicolumn{3}{c}{78.68\%} & \multicolumn{3}{c}{86.58\%} & \multicolumn{3}{c}{80.77\%}\\
             \multicolumn{2}{l}{4-bit DGMS (w/o transfer)} & \multicolumn{3}{c}{70.25\%} & \multicolumn{3}{c}{77.90\%} & \multicolumn{3}{c}{86.39\%} & \multicolumn{3}{c}{80.41\%}\\
             \midrule[0.6pt]
             \multirow{3}*{\shortstack{4-bit DGMS\\ (w/ transfer)}}
             & Source & CUB & Cars & Air & Img & Cars & Air & Img & CUB & Air & Img & CUB & Cars\\
             \cmidrule(lr){2-2}
             \cmidrule(lr){3-5} \cmidrule(lr){6-8}
             \cmidrule(lr){9-11}
             \cmidrule(lr){12-14}
              & \cellcolor{linecolor1}\scshape{Zero-Shot} & \cellcolor{linecolor1}34.69\% & \cellcolor{linecolor1}62.31\% & \cellcolor{linecolor1}35.09\% & \cellcolor{linecolor1}73.53\% & \cellcolor{linecolor1}74.29\% & \cellcolor{linecolor1}66.44\% & \cellcolor{linecolor1}82.28\% & \cellcolor{linecolor1}81.46\% & \cellcolor{linecolor1}71.75\% & \cellcolor{linecolor1}77.46\% & \cellcolor{linecolor1}74.97\% & \cellcolor{linecolor1}77.31\%\\
             & \cellcolor{linecolor}\scshape{One-Epoch} & \cellcolor{linecolor}68.37\% & \cellcolor{linecolor}69.13\% & \cellcolor{linecolor}68.80\% & \cellcolor{linecolor}77.70\% & \cellcolor{linecolor}77.54\% & \cellcolor{linecolor}77.50\% & \cellcolor{linecolor}85.70\% & \cellcolor{linecolor}85.84\% & \cellcolor{linecolor}85.79\% & \cellcolor{linecolor}79.90\% & \cellcolor{linecolor}79.87\% & \cellcolor{linecolor}80.14\%\\
        \bottomrule[0.95pt]
        \end{tabular}
        }
\end{table*}
\subsection{Gaussian Mixture Evolution}
For DGMS, the quantization mapping is adaptively learned with the GM evolution during training.
In order to further understand this weight-distribution co-tuning schema, as shown in Fig.~\ref{fig:distribution}, we visualize the weight redistribution process of 2-bit quantization (\ie, $K=3$) with DGMS and the learning curve of ResNet-20 on CIFAR-10.
Fig.~\ref{fig:distribution}(a) shows the pretrained  weight distribution with the initial GM densities plotted.
Fig.~\ref{fig:distribution}(b) and (c) respectively illustrate the distributions of weights after DGMS and quantization.
We can observe that the proposed DGMS is capable of tuning the GM according to the continuous distribution. Besides, the model-inherent statistical information is modeled and the weights are accordingly redistributed.
Furthermore, the weights projected by DGMS serve as a fine hard-quantized estimation, which manages to narrow the gap between the original and compressed representations.
Fig.~\ref{fig:distribution}(d) plots the validation performance and the weight deviation measured by mean square error (MSE) of the uncompressed and quantized weights. 
Note that no such quantization error is used for DGMS training, but it can be seen that the quantization error spontaneously declines with accuracy boosting.
\subsection{Domain Invariance}
In Tab.~\ref{tab:domain_study}, we evaluate the transfer ability with 4-bit ResNet-18 across ImageNet, CUB200-2011~\cite{CUB200}, Stanford Cars~\cite{Cars}, and FGVC Aircraft~\cite{Aircraft}.
We directly transfer the parameterized sub-distribution from the source domain to perform DGMS quantization and report the Top-1 accuracy (\%) without any or with only \textit{one-epoch} tuning.
It is noteworthy that the zero epoch results are \textit{zero-shot} quantization results on the target dataset via domain transfer.
From the table, we can observe that except for the large-scale ImageNet dataset, all the models achieve competitive performances compared to full-precision models with zero training.
This is because the sub-distribution learning has converged enough for the three medium-scale datasets, but it is not adapted to ImageNet with a reparable gap.
Therefore, after only one epoch of fine-tuning with DGMS, the models are able to match the performance of uncompressed or vanilla DGMS-quantized models (trained for 60 epochs).
This demonstrates the domain-invariant character of the sub-distribution, which is a model-inherent representation with superior generalization performance and training efficiency. 
Beyond that, this also provides a novel angle orthogonal to existing methods for zero-shot model quantization
, \textit{i.e.}, through domain generalization of adaptive quantization parameters.
\section{Conclusions}
In this paper, we introduce a novel quantization method named DGMS to automatically find a latent task-optimal low-bit sub-distribution, forming a distributional bridge between full-precision and quantized representations.
Weights are projected based on the GM-parameterized sub-distribution approximation, which evolves with weights in a joint fashion by directly optimizing the task-objective. 
As a result, DGMS adaptively quantizes the DNNs with a found task-optimal sub-distribution and achieves negligible performance loss with high compression rates.
Extensive experiments on both image classification and detection on over-parameterized and lightweight DNNs have been conducted, demonstrating a competitive compression performance compared to other state-of-the-art methods and the transferability of the obtained model-inherent sub-distributions across different domains. 
To date, the formulation and optimization direction remain open problems in the quantization community. Among numerous existing solutions, we propose to view the optimal quantization configuration searching in a different way as to find the task-optimal sub-distribution. 
We also provide a great potential to solve zero-shot quantization through domain generalization, and we hope our work could spur future related researches.
In the future, we would like to explore DGMS with a bimodal distribution (\eg, arcsine distribution) for binary DNNs, and it would be intriguing to solve the task-optimal sub-distribution searching problem with techniques like reinforcement learning. 


\bibliography{main}
\bibliographystyle{icml2022}

\newpage
\appendix
\onecolumn
\section{Details of Q-SIMD}\label{sec:QSIMD}
\begin{figure*}[ht]
    \centering
    \includegraphics[width=5.20in]{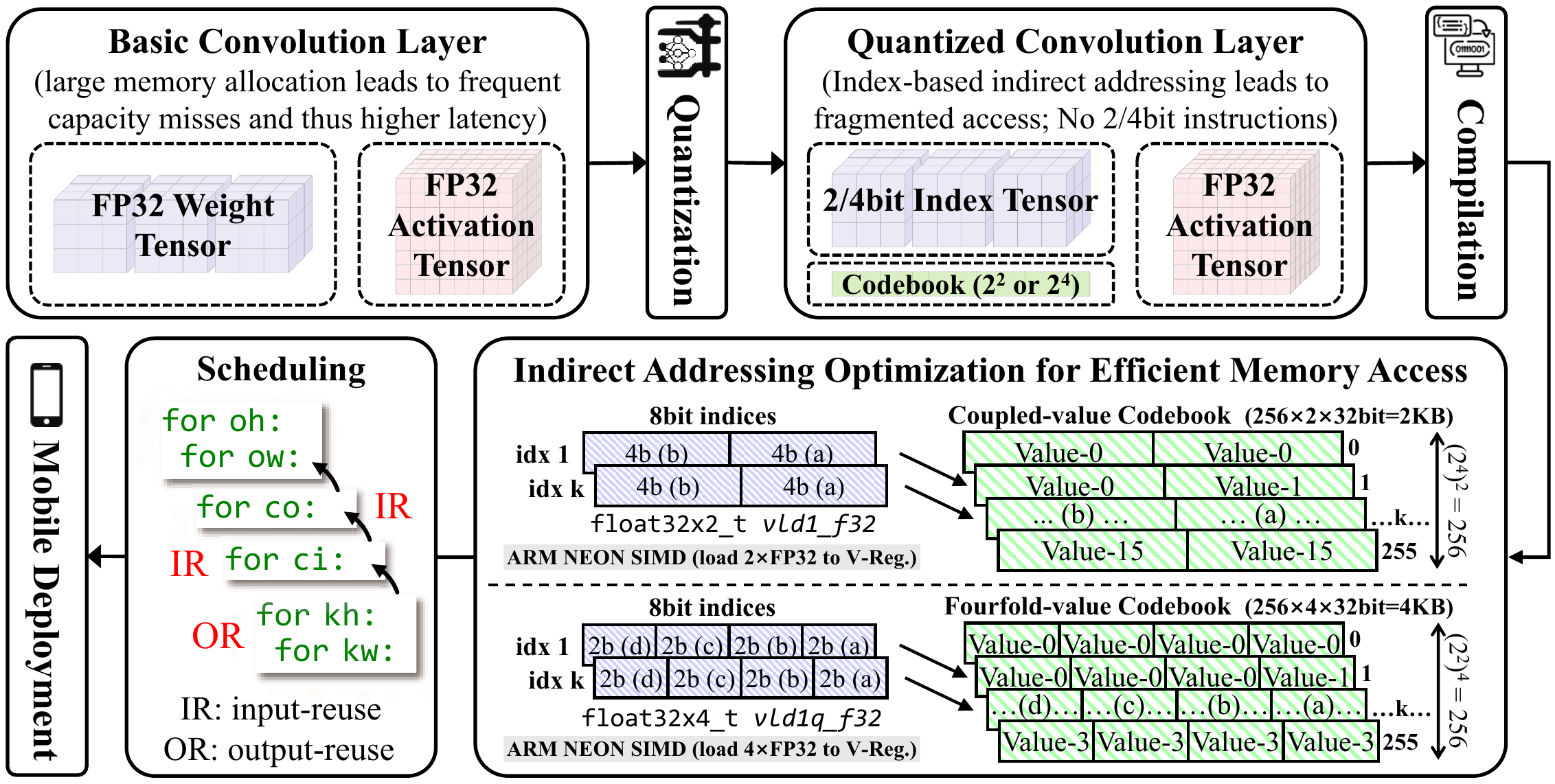}
    \caption{Overview of Q-SIMD compilation framework for weight sharing quantization on the mobile SIMD CPU. The scheduling also employs loop tiling and unrolling (not shown in the figure). \texttt{kw}, \texttt{kh}: width and height of kernel. \texttt{ci}, \texttt{co}: input and output channel. \texttt{ow}, \texttt{oh}: width and height of output feature map.}
    \label{fig:compiler}
\end{figure*}
Fig.~\ref{fig:compiler} shows the detailed illustration of our Q-SIMD flow.
DGMS firstly quantizes a full-precision model to a low-bit model of 2-bit or 4-bit weights. 
Because the real values of weights are still FP32, the compact model is stored as a low-bit index tensor with the corresponding codebook. 
Naturally, when executing the multiply-add operations, the real weights are loaded through indirect addressing with indices. 
However, the current ISA does not support 4-bit or 2-bit operations and requires at least 8-bit alignment.
Additionally, the indirect addressing is irregular and thus damages the SIMD operations.

To address this problem mentioned above, we introduce an extended codebook with coupled or fourfold values for 4-bit and 2-bit quantization, respectively.
In this way, the number of codebook entries is increased to 256, and then the index bit-width is changed to be 8-bit.
With the ARM NEON SIMD instructions (\eg, \texttt{float32x2\_t vld1\_f32} and \texttt{float32x4\_t vld1q\_f32}), we can avoid bit wasting for low-bit alignment and maintain the parallel data loading for SIMD operations.

Furthermore, given that we mainly focus on weight quantization, the bottleneck of memory access turns to be activations.
Therefore, we reduce the activation reloading via output-reuse (OR) and input-reuse (IR) scheduling.
For the inner-loops, we keep the outputs local in the L1 cache to avoid reloading the partial sums.
For the outer-loops, we put the \texttt{oh} and \texttt{ow} dimensions to reuse inputs across different output channels.

The aforementioned operation is registered as a customized operator in TVM~\cite{TVM}. With the Android TVM RPC tool, the quantized models can be conveniently deployed on the ARM CPU and evaluated for inference time.

\section{Further Discussions}
\subsection{Sub-Distribution}\label{sec:sub-ditribution}
In this section, we discuss about sub-distribution introduced in Definition~\ref{def:subdistribution}, which is essential and serves for the moral of this paper.
\subsubsection{Statistical Approximation} 
The concept ``approximate equivalence'', denoted as $\approxeq$, is statistical and it means the divergence between the estimated and the real data distribution does not exceed a small error $\epsilon$.
\subsubsection{Full-Precision Distribution Estimation} 
The full-precision distribution estimation is the first condition in Definition~\ref{def:subdistribution}, \ie, $\mathcal{D}_{\mathcal{S}} \approxeq \mathcal{D}_{\mathcal{W}}$. 
As stated before, the GM sub-distribution is initialized by MLE method, and MLE can model the statistical feature of data samples based on a parameterized approximation~\cite{EM}. 
Hence, the parameterized GM is approximately equivalent to the data's real distribution which can not be precisely measured but can be statistically estimated~\cite{GMM}.
\subsubsection{Categorical Distribution Approaching}
The quantization distribution approaching is the second condition in Definition~\ref{def:subdistribution}, \ie, $\mathop{\lim}\limits_{\tau \rightarrow 0} \mathcal{D}_{\mathcal{S}} \approxeq \mathcal{D}_{\mathcal{Q}}$.
This means after hard quantization, the weights are re-distributed to discrete values according to the categorical distribution $\mathcal{D}_{\mathcal{Q}}$ that originates from the concrete GM. 
Formally, given full-precision weight $\mathbf{w} \sim \mathcal{D}_{\mathcal{W}}$, $\Phi(\mathbf{w}) \sim \mathcal{D}_{\mathcal{S}}$ as the statistically shared weights according to the GM model and $\Psi(\mathbf{w}) \sim \mathcal{D}_{\mathcal{Q}}$ as the quantized value. 
Assuming that $k = \mathop{\arg\max}\limits_{i} \varphi(\mathbf{w}, i)$, then we have $\lim\limits_{\tau \rightarrow 0}\phi_k(\mathbf{w})=1$ and $\lim\limits_{\tau \rightarrow 0}\phi_{j}(\mathbf{w})=0\vert_{j\not = k}$, hence it can be easily proved that $\lim\limits_{\tau \rightarrow 0}\Phi(\mathbf{w})= \Psi(\mathbf{w})$ (notations taken from Sec.~\ref{sec:DGMS}).
\subsection{MDL View on DGMS}\label{sec:mdl}
Supposing that we have the training dataset 
$\mathcal{D} = \{\mathcal{X}, \mathcal{Y}\} = \{\mathbf{x}_i, \mathbf{y}_i\}_{i=1}^{N}$ and 
neural network $\mathcal{F}$ with 
weights $\mathcal{W}$, DGMS directly optimizes the objective with gradient descent:
\begin{equation}\label{equ:DGMSopt}
    \min_{\mathcal{W}, \vartheta, \tau} 
    \mathcal{L}^E= -\log p\Big(\mathcal{Y}\vert\mathcal{X}, \Phi\big(\mathcal{W};\vartheta, \tau\big)\Big),
\end{equation}
where the Gaussian Mixture is parameterized by 
$\vartheta=\{\pi_k, \mu_k, \gamma_k\}_{k=1}^{K}$ and weights are
smoothly quantized by the differentiable indicator
defined in Eqn. (\ref{equ_softmask}).

From the minimum description length (MDL) view in information theory~\cite{Shannon, MDL}, $\mathcal{L}^E$ is the lower bound of the information cost expectation to fit the data~\cite{MDL1, MDL2}. 
In contrast, the previous works~\cite{GumbelSoftmax, ConcreteDistribution, SWS, DPBNN} optimize the \textit{evidence lower bound} (ELBO) to train DNNs with smoothed stochastic weights from an estimated categorical prior distribution (\textit{e.g.} Gaussian Mixture).
The objective can be written with the error cost $\mathcal{L}^{E}$ plus the complexity cost $\mathcal{L}^{C}$, and for DGMS this can be given as follows:
\begin{equation}\label{equ:Bayesianopt}
    \min_{\mathcal{W}, \vartheta, \tau}
    \underbrace{-\log p \Big(\mathcal{Y}\vert\mathcal{X}, 
    \Phi \big(\mathcal{W};\vartheta\big)\Big)}_{\mathcal{L}^E} +
    \underbrace{-\log p\Big(\Phi\big(\mathcal{W},\vartheta,\tau\big)\Big) +
    \mathcal{H}\Big(q_{\vartheta}\big(\mathcal{W}\big)\Big)}_{\mathcal{L}^C},
\end{equation}
where $\mathcal{L}^C$ amounts to the Kullback-Leibler divergence between the prior distribution and the Bayesian posterior distribution $q_{\vartheta}(\mathcal{W})$, and $\mathcal{H}$ is the entropy term which can be approximately reduced to a constant~\cite{SWS}.
In this paper, the complexity cost $\mathcal{L}^C$ is dropped while the weights and the parameterized GM co-adapt for redistribution.

\subsection{Hardware Deployment}
\subsubsection{Limitations of codebook-based fully-adaptive quantization}
The direct quantization (\textit{e.g.}, 4-bit uniform quantization) can avoid the indirect addressing. However, this kind of quantization is not accuracy-friendly. In this work, even though the codebook takes up the memory footprint, the overhead is fairly small because our quantization granularity is layer-wise. Thanks to the small memory footprint and the cache locality, the codebook can be held in the L1-cache whose access latency is fairly low. Besides, the codebook-based quantization may lead to irregular memory access from indirect addressing. To solve this problem, we introduce the index-pair to access data in parallel (see Q-SIMD depolyment paradigm in Sec.~\ref{sec:QSIMD}).
\subsubsection{Detailed memory footprint of codebooks}
The memory footprint of parameters in a quantized layer (\textit{e.g.}, 4-bit) is calculated by: \#params$\times$4 (bit). The codebook overhead is calculated by: $2^4\times 32$ (bit). For example, in the 4-bit quantized ResNet-18, ResNet-50, MnasNet-A1, SSDLite-MBV2, and ProxylessNAS, the overhead ratios of codebook are: 0.02\%, 0.03\%, 0.2\%, 0.3\%, and 0.3\%.

\section{Details of Experiments}\label{app:imp}
\subsection{Datasets}
\paragraph{CIFAR-10}
The CIFAR-10 dataset~\cite{CIFAR} consists of 60,000 32$\times$32 tiny images for 10 classes object recognition, with 6,000 images per class.
\paragraph{ImageNet} The ImageNet (ILSVRC 2012) dataset~\cite{ImageNet} is a challenging large-scale dataset for image classification, containing 1.3M training samples and 50K test images with 1,000 object classes for recognition.
\paragraph{PASCAL VOC}
The PASCAL VOC dataset~\cite{PASCAL, PASCAL07} is widely used as both semantic segmentation and object detection benchmarks.
For object detection, VOC2007 \texttt{trainval} and VOC2012 \texttt{trainval} respectively contain 5,011 images and 11,540 images for training (16,551 images in all) and VOC07 \texttt{test} contains 4,952 images for evaluation.
PASCAL VOC involves 21 classes including the a background class for detection and segmentation.
\paragraph{CUB200-2011}
The CUB200-2011 dataset~\cite{CUB200} is a medium-scale dataset for fine-grained classification task, which contains 11,788 images in all for 200 bird species (5,994 samples for training and 5,794 images for validation). 
\paragraph{Stanford Cars}
The Stanford Cars dataset~\cite{Cars} contains a total of 16,185 car images of 196 categories, of which 8,144 samples are used for training and 8,041 images are used for validation.
\paragraph{FGVC Aircraft}
The FGVC Aircraft dataset~\cite{Aircraft} consists of 10,000 images for 100 aircraft variants, the training set involves 6,667 samples and validation set involves 3,333 images. 
\subsection{Implementation Details}
\paragraph{Optimization}
We set the batch size as $128$ on CIFAR-10 and $256$ on ImageNet for all the models. SGD with $0.9$ momentum and $5\times10^{-4}$ weight-decay is used during training. All the models are trained for 350 epochs and 60 epochs respectively on CIFAR-10 and ImageNet.
The max learning rate is set to $2\times10^{-5}$ ($1\times10^{-5}$ for 2-bit ResNet-50) using one-cycle scheduler~\cite{OneCycle} and the initial temperature $\tau$ in Eqn. (\ref{equ:gumbel}) is $0.01$  for all the experiments.
The training configurations for transferability experiments on the medium-scale datasets (CUB200-2011, Stanford Cars and FGVC Aircraft) are the same as ImageNet. 
For PASCAL VOC detection, the batch size is 32 with $1\times10^{-5}$ learning rate and we employ the cosine scheduler. The detectors are trained for only 10 epochs.
With regard to the data augmentation, we simply adopt random crop and horizontal flip for image classification and the same augmentation operations as that in public source\footnote{\url{https://github.com/qfgaohao/pytorch-ssd}.} for object detection.
\paragraph{Implementation}
The experiments are implemented with PyTorch 1.6~\cite{Pytorch} on PH402 SKU 200 with 32G memory GPU devices.
All the layers except the first and the last are quantized, which has become a common practice for model quantization.
Similarly for detectors, we perform the quantization on all layers except for the first layer of the model and the last layer of the classification head and regression head.
\begin{table*}[ht]
    \centering
    \caption{Ablation study on initialization. The results are Top-1 classification accuracy (\%) on CIFAR-10.}
    \label{tab:inits}
    \begin{tabular}{lccccc}
    \toprule[0.95pt]
    \multirow{2}{*}[-0.5ex]{Model} &\multirow{2}{*}[-0.5ex]{Bits}& \multirow{2}{*}[-0.5ex]{Std. Init.} &
    \multicolumn{3}{c}{Empirical Initialization}\\
    \cmidrule(lr){4-6}
    & & & $\gamma=$0.010 & $\gamma=$0.005 & $\gamma=$0.001\\
    \midrule[0.6pt]
    \multirow{2}{*}[-0.5ex]{ResNet-20} & 2 & 89.26\% &92.13\% & 90.88\% & 91.30\%\\
    & 3 & 92.16\% & 92.84\% & 92.54\% & 92.34\%\\
    \midrule[0.6pt]
    \multirow{2}{*}[-0.5ex]{VGG-Small} & 2 & 92.55\% & 94.36\% & 94.40\% & 94.41\%\\
    & 3 & 94.30\% & 94.46\% & 94.56\% & 94.52\%\\
    \bottomrule[0.95pt]
    \end{tabular}
\end{table*}
\section{Further Ablation Study}
\subsubsection{Initialization}
For the GM initialization, we adopt k-means which can be simply and efficiently implemented on GPU devices\footnote{\url{https://github.com/subhadarship/kmeans_pytorch}.}.
In the CIFAR-10 experiments, we test two types of initialization for the parameter $\gamma$, including standard deviation initialization and empirical initialization.
As a result, we empirically find that the calculation of standard deviation $\gamma$ in Algorithm~\ref{alg:alg} may not be optimal but is simple and effective compared to the empirical initialization (see Tab.~\ref{tab:inits}).
\subsubsection{Temperature}
\begin{wraptable}{r}{0.5\linewidth}
    \vspace{-12.75pt}
    \centering
    \caption{Ablation study on temperature hyper-parameter $\tau$. The results are Top-1 classification accuracy (\%) with 3-bit VGG-Small on CIFAR-10.}
    \label{tab:temp}
    \begin{tabular}{lccc}
    \toprule[0.95pt]
    \multirow{2}{*}[-0.5ex]{Temperature Type} & \multicolumn{3}{c}{Initialization Value}\\
    \cmidrule(lr){2-4}
    & $\tau$=0.100 & $\tau$=0.010 & $\tau$=0.001\\
    \midrule[0.6pt]
    Fixed & 93.57\% & 94.56\% & 94.27\%\\
    Learned & 93.54\% & 94.64\%& 94.04\%\\
    \bottomrule[0.95pt]
    \end{tabular}
\end{wraptable}
As has been stated in Sec.~\ref{sec:DGMS}, the temperature hyper-parameter $\tau$ controls the discretization estimation level.
And when $\tau$ approaches zero, the estimated representation is approximately equivalent to the quantized one.
This hyper-parameter can be set as a fixed or learnable parameter with a predefined initialized value.
In Tab.~\ref{tab:temp}, we test 3-bit VGG-Small on CIFAR-10 with different initializations.
One can observed that a relatively lower temperature is more recommended to bridge the quantization gap, and a fixed temperature can achieve competitive performances compared to an adaptively learned one.

\newcommand{\coo}{\ensuremath{\mathrm{CO_2}}}
\section{Broader Impact}
The rapid development of deep learning brings huge convenience and advanced productivity to our society, together with a higher daily cost (\textit{e.g.}, carbon footprint).
The proposed DGMS is a novel quantization method to address this issue, and it can be utilized to compress a variety of DNNs.
In real-life, this technique provides an eco-friendly and energy-efficient deployment solution for deep learning models, profiting from lower power consumption and less $\coo$ emission.
Though DGMS facilitates the AI development on edge devices like mobile phones, if without legal restraint, it will potentially lead to the infringement of image rights and personal privacy.

\end{document}